\documentclass{article}
\usepackage{spconf,amsmath,graphicx}
\usepackage[hyphens]{url}
\usepackage{amsmath}
\usepackage{booktabs}
\usepackage[dvipsnames]{xcolor}
\RequirePackage{xcolor}
\usepackage{multirow}
\usepackage[hyphens]{url}

\def\ie{{\em i.e}}
\def\eg{{\em e.g}}
\def\etal{{\em et al. }}
\def\eg{\emph{e.g}}
\def\etal{\emph{et al. }}
\def\ie{\emph{i.e}}
\def\t{\vec{t}}

\begin{document}
\title{Exposing Deep Fakes Using Inconsistent Head Poses}

\name{Xin Yang$^\star$, Yuezun Li$^\star$ and Siwei Lyu\thanks{$^\star$ The authors contribute equally.}}
\address{University at Albany, State University of New York, USA}
\maketitle

\begin{abstract}
In this paper, we propose a new method to expose AI-generated fake face images or videos (commonly known as the {\em Deep Fakes}). Our method is based on the observations that Deep Fakes are created by splicing synthesized face region into the original image, and in doing so, introducing errors that can be revealed when 3D head poses are estimated from the face images. We perform experiments to demonstrate this phenomenon and further develop a classification method based on this cue. Using features based on this cue, an SVM classifier is evaluated using a set of real face images and Deep Fakes.

\begin{keywords}
Media Forensics, DeepFake Detection, Head Pose Estimation
\end{keywords} 

\end{abstract}

\section{Introduction}

\begin{figure*}[t]
	\centering
	\includegraphics[width=.95\linewidth]{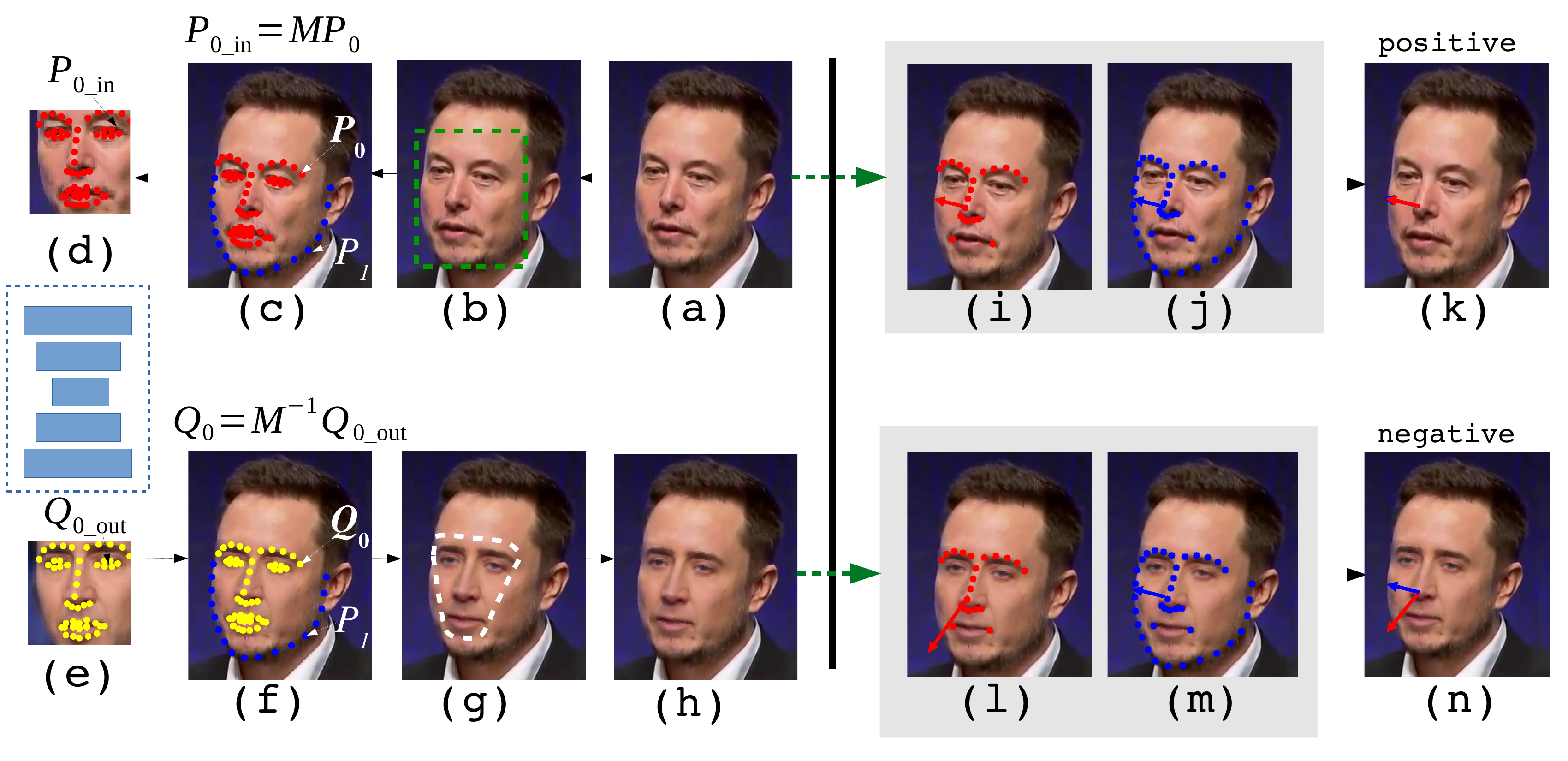}
	~\vspace{-1em}
	\caption{\em  \small Overview of Deep Fake work-flow (Left) and our method (Right). In {\bf(Deep Fake work-flow)}: {\bf(a)} is the original image. {\bf(b)} Detected face in the image. {\bf(c)} Detected 2D facial landmarks. {\bf(d)} Cropped face in (a) is warped to a standardized face using an affine transformation $M$. {\bf(e)} Deep Fake face synthesized by the deep neural network. {\bf(f)} Deep Fake face is transformed back using $M^{-1}$. {\bf(g)} The mask of transformed face is refined based on landmarks. {\bf(g)} Synthesized face is merged into the original image. {\bf(h)} The final fake image. For {\bf(our method)}: The top row corresponds to a real image and the bottom corresponds to a Deep Fake. We compare head poses estimated using facial landmarks from the whole face {\bf (j)},{\bf (m)} or only the central face region {\bf (i)}, {\bf (l)}. The alignment error is revealed as differences in the head poses shown as their projections on the image plane. The difference of the head poses is then fed to an SVM classifier to differentiate the original image {\bf (k)} from the Deep Fake {\bf (n)}.}
	\label{fig:overview&method}
	~\vspace{-2em}
\end{figure*}

Thanks to the recent developments of machine learning, the technologies for manipulating and fabricating images and videos have reached a new level of sophistication. The cutting edge of this trend are the so-called Deep Fakes, which are created by inserting faces synthesized using deep neural networks into original images/videos. Together with other forms of misinformation shared through the digital social network, Deep Fakes created digital impersonations have become a serious problem with negative social impact \cite{chesney_citron_2018}. Accordingly, there is an urgent need for effective methods to expose Deep Fakes. 

To date, detection methods of Deep Fakes have relied on  artifacts or inconsistencies intrinsic to the synthesis algorithms, for instance, the lack of realistic eye blinking \cite{li2018ictu} and mismatched color profiles \cite{li2018detection}. Neural network based classification approach has also been used to directly discern real imagery from Deep Fakes \cite{afchar2018mesonet}. In this work, we propose a new approach to detect Deep Fakes. Our method is based on an intrinsic limitations in the deep neural network face synthesis models, which is the core component of the Deep Fake production pipeline. Specifically, these algorithms create faces of a different person but keeping the facial expression of the original person. However, the two faces have mismatched facial landmarks, which are locations on human faces corresponding to important structures such as eye and mouth tips, as the neural network synthesis algorithm does not guarantee the original face and the synthesized face to have consistent facial landmarks, as shown in Fig. \ref{fig:overview&method}.  



{The errors in landmark locations may not be visible directly to human eyes, but can be revealed from head poses (\ie, head orientation and position) estimated from 2D landmarks in the real and faked parts of the face. Specifically, we compare head poses estimated using all facial landmarks and those estimated using only the central region, as shown in Fig. \ref{fig:overview&method}. The rationale is that the two estimated head poses will be close for the original face, Fig. \ref{fig:overview&method}(k). But for a Deep Fake, as the central face region is from the synthesized face, the errors due to the mismatch of landmark locations from original and generated images aforementioned will lead to a larger difference between the two estimated head poses, Fig. \ref{fig:overview&method}(n).  We experimentally confirm the significant difference in the estimated head pose in Deep Fakes. Then we use the difference in estimated head pose as a feature vector to train a simple SVM based classifier to differentiate original and Deep Fakes. Experiments on realistic Deep Fake videos demonstrate the effectiveness of our algorithm. }

\section{\bf Deep Fake Production Pipeline}

{
The overall process of making Deep Fakes is illustrated in  Fig. \ref{fig:overview&method}(a)-(h). To generate a Deep Fake, we feed the algorithm an image (or a frame from a video) that contains the source face to be replaced. Bounding box of faces are obtained with a face detector, followed by the detection of facial landmarks. The face area is warped into a standard configuration through affine transformation $M$, by minimizing the alignment errors of central facial landmarks (red dots in Fig. \ref{fig:overview&method}(c)) to a set of standard landmark locations, a process known as face alignment. This image is then cropped into $64 \times 64$ pixels, and fed into the deep generative neural network to create the synthesized face. The synthesized face is transformed back with $M^{-1}$ to match the original face.  Finally, with post-processing such as boundary smoothing, a Deep Fake image/video frame is created.}

\section{3D Head Pose Estimation}\label{headpose_estimation}

\begin{figure}[t]
	\centering
	\includegraphics[width=.55\linewidth]{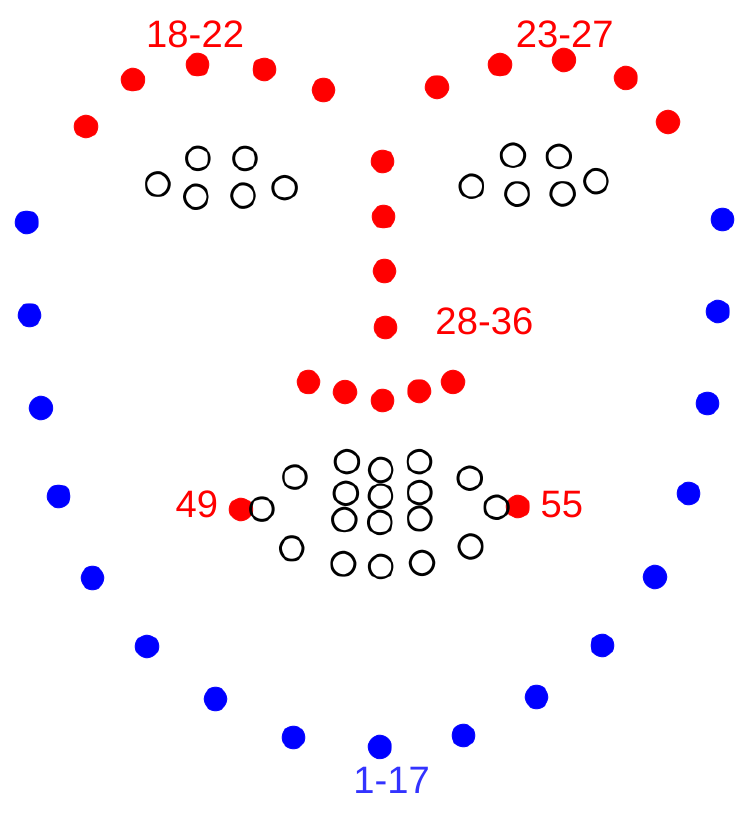}
	~\vspace{-1em}
	\caption{\em \small $68$ facial landmarks. {Red dots are used as central face region. Blue and red landmarks are used as whole face. The landmarks represented as empty circles are not used in head pose estimation.}}
	\label{fig:facial_landmarks}
	~\vspace{-3em}
\end{figure}

The 3D head pose corresponds to the rotation and translation of the world coordinates to the corresponding camera coordinates. Specifically, denote $[U, V, W]^T$ as the world coordinate of one facial landmark, $[X, Y, Z]^T$ be its camera coordinates, and $(x, y)^T$ be its image coordinates. The transformation between the world and the camera coordinate system can be formulated as
\begin{equation}
\left[{\begin{array}{c} X \\ Y \\ Z \end{array}}\right] = 
R
\left[{\begin{array}{c} U \\ V \\ W \end{array}}\right] + \t,
\label{equ:rotate_translate}
\end{equation}
where $R$ is the $3 \times 3$ rotation matrix, $\t$ is $3 \times 1$ translation vector. The transformation between camera and image coordinate system is defined as
\begin{equation}
s\left[{\begin{array}{c} x \\ y \\ 1 \end{array}}\right] = 
\left[{\begin{array}{ccc} f_x & 0 & c_x \\ 0 & f_y & c_y \\ 0 & 0 & 1 \end{array}}\right]
\left[{\begin{array}{c} X \\ Y \\ Z \end{array}}\right]
\label{equ:project}
\end{equation}
where $f_x$ and $f_y$ are the focal lengths in the $x$- and $y$-directions and $(c_x, c_y)$ is the optical center, and $s$ is an unknown scaling factor.

In 3D head pose estimation, we need to solve the reverse problem, \ie, estimating $s$, $R$ and $\t$ using the 2D image coordinates and 3D world coordinates of the same set of facial landmarks obtained from a standard model, \eg, a 3D avearge face model, assuming we know the camera parameter. Specifically, for a set of $n$ facial landmark points, this can be formulated as an optimization problem, as
\[
\min_{R,\t,s} \sum_{i=1}^n \left\|s\left[{\begin{array}{c} x_i \\ y_i \\ 1 \end{array}}\right] - \left[{\begin{array}{ccc} f_x & 0 & c_x \\ 0 & f_y & c_y \\ 0 & 0 & 1 \end{array}}\right]\left(R
\left[{\begin{array}{c} U_i \\ V_i \\ W_i \end{array}}\right] + \t\right) \right\|^2
\]
that can be solved efficiently using the Levenberg-Marquardt algorithm \cite{opencv_library}. The estimated $R$ is the camera pose which is the rotation of the camera with regards to the world coordinate, and the head pose is obtained by reversing it as $R^T$ (as $R$ is an orthornormal matrix).  

\section{Inconsistent Head Poses in Deep Fakes}\label{inconsistency}

{
As a result of swapping faces in the central face region in the Deep Fake process in Fig. \ref{fig:overview&method}, the landmark locations of fake faces often deviate from those of the original faces. As shown in Fig. \ref{fig:overview&method}(c), a landmark in the central face region $P_{0}$ is firstly affine-transformed into $P_{0\_in} = MP_{0}$. After the generative neural network, its corresponding landmark on the faked face is $Q_{0\_out}$. 

As the configuration of the generative neural network in Deep Fake does not guarantee landmark matching, and people have different facial structures, this landmark $Q_{0\_out}$ on generated face could have different locations to $P_{0\_in}$. Based on the comparison 51 central region landmarks of 795 pairs of images in 64 $\times$ 64 pixels, the mean shifting of a landmark from the input (Fig. \ref{fig:overview&method}(d)) to the output (Fig. \ref{fig:overview&method}(e)) of the generative neural network is 1.540 pixels, and its standard deviation is 0.921 pixels. After an in versed transformation $Q_{0} = M^{-1}Q_{0\_out}$, the landmark locations $Q_{0}$ in the faked faces will differ from the corresponding landmarks $P_{0}$ in the original face. However, due to the fact that Deep Fake only swap faces in the central face region, the locations of the landmarks on the outer contour of the face (blue landmarks $P_{1}$ in Fig. \ref{fig:overview&method}(c) and (f)) will remain the same. This mismatch between the landmarks at center and outer contour of faked faces is revealed as inconsistent 3D head poses estimated from central and whole facial landmarks. Particularly, the head pose difference between central and whole face region will be small in real images, but large in fake images.
}

\begin{figure}[t]
	\centering
	\includegraphics[width=.95\linewidth]{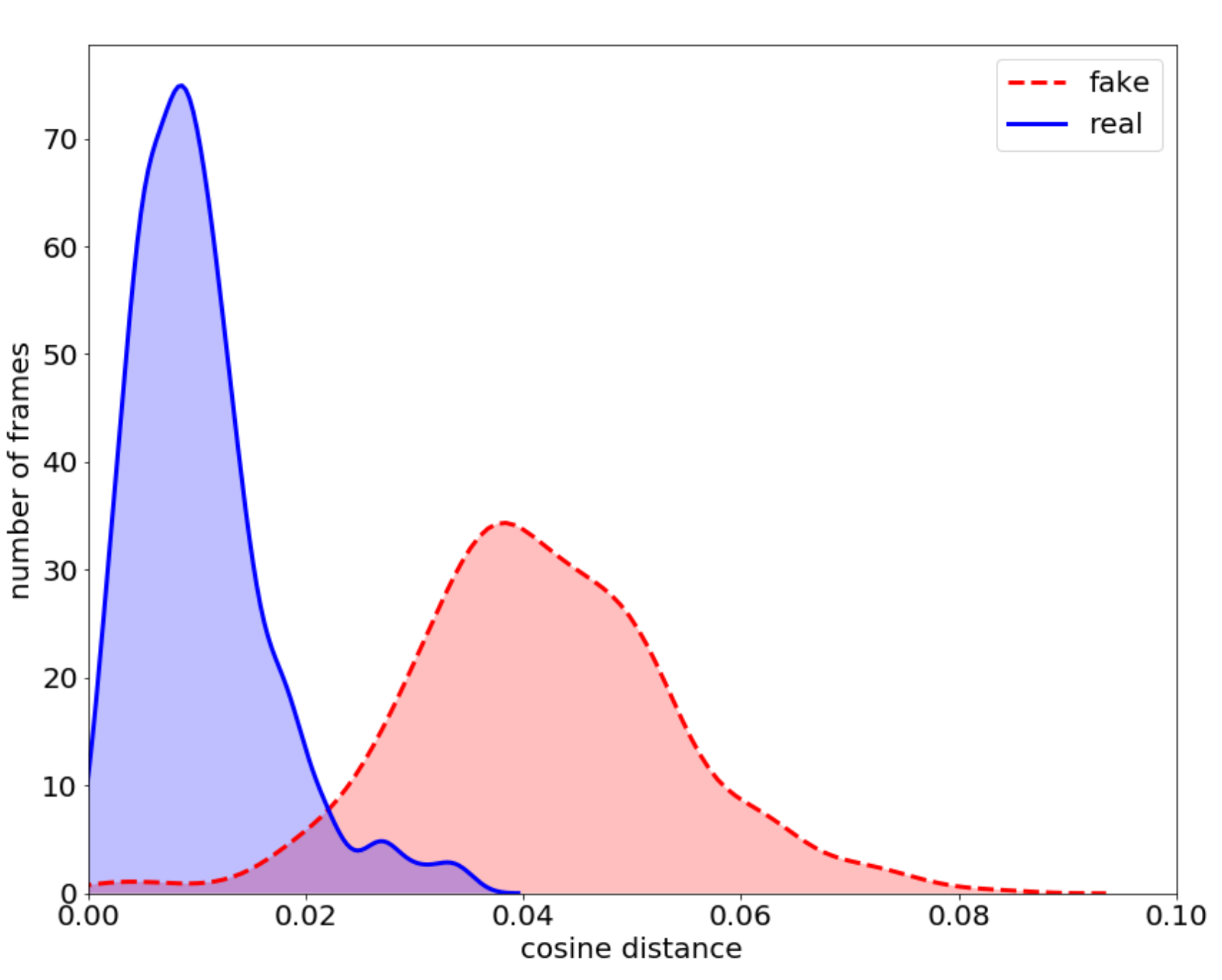}
	~\vspace{-1em}
	\caption{\em \small Distribution of the cosine distance between $\vec{v}_c$ and $\vec{v}_a$ for fake and real face images.}
	\label{fig:AC_difference}
	~\vspace{-2em}
\end{figure}

We conduct experiments to confirm our hypothesis. For simplicity, we look at the head orientation vector only. Denote $R_a^T$ as the rotation matrix estimated using facial landmarks from the whole face (red and blue landmarks in Fig. \ref{fig:facial_landmarks}) using method described in Section \ref{headpose_estimation}, and $R_c^T$ as the one estimated using only landmarks in the central region (red landmarks in Fig. \ref{fig:facial_landmarks}). We obtain the 3D unit vectors $\vec{v}_a$ and $\vec{v}_c$ corresponding to the orientations of the head estimated this way, as $\vec{v}_a = R_a^T \vec{w}$ and $\vec{v}_c = R_c^T \vec{w}$, respectively, with $\vec{w} = [0,0,1]^T$ being the direction of the $w$-axis in the world coordinate.  We then compare the cosine distance between the two unit vectors $\vec{v}_c$ and $\vec{v}_a$, $1-\vec{v}_a \cdot \vec{v}_c / (\| \vec{v}_a \| \| \vec{v}_c \|)$, which takes value in $[0,2]$ with $0$ meaning the two vectors agree with each other. The smaller this value is, the closer the two vectors are to each other.  Shown in Fig. \ref{fig:AC_difference} are histograms of the cosine distances between $\vec{v}_c$ and $\vec{v}_a$ for a set of original and Deep Fake generated images. As these results show, the cosine distances of the two estimated head pose vectors for the real images concentrates on a significantly smaller range of values up to $0.02$, while for Deep Fakes the majority of the values are in the range between $0.02$ and $0.08$. The difference in the distribution of the cosine distances of the two head orientation vectors for real and Deep Fakes suggest that they can be differentiated based on this cue.

\section{Classification based on Head Poses}

We further trained SVM classifiers based on the differences between head poses estimated using the full set of facial landmarks and those in the central face regions to differentiate Deep Fakes from real images or videos. The features are extracted in following procedures: (1) For each image or video frame, we run a face detector and extract $68$ facial landmarks using software package {\tt DLib} \cite{dlib09}. (2) Then, with the standard 3D facial landmark models of the same $68$ points from OpenFace2 \cite{baltrusaitis2018openface}, the head poses from central face region ($R_c$ and $t_c$) and whole face ($R_a$ and $t_a$) are estimated with landmarks $18-36, 49, 55$ (red in Fig. \ref{fig:facial_landmarks}) and $1-36, 49, 55$ (red and blue in Fig. \ref{fig:facial_landmarks}), respectively. Here, we approximate the camera focal length as the image width, camera center as image center, and ignore the effect of lens distortion. (3) The differences between the obtained rotation matrices $(R_a - R_c)$ and translation vectors $(\vec{t}_a - \vec{t}_c)$ are flattened into a vector, which is standardized by subtracting its mean and divided by its standard deviation for classification.


The training and testing data for the SVM classifier are based on two datasets of real and Deep Fake images and videos. The first, {\tt UADFV}, is a set of Deep Fake videos and their corresponding real videos that are used in our previous work \cite{li2018ictu}. This dataset contains $49$ real videos, which were used to create $49$ Deep Fake videos.   The average length of these videos is approximately $11.14$ seconds, with a typical resolution of $294 \times 500$ pixels. The second data set is a subset from the DARPA MediFor GAN Image/Video Challenge 
, which has $241$ real images and $252$ Deep Fake images. For the training of the SVM classifier, we use frames from 35 real and 35 Deep Fake videos in the UADFV dataset, with a total number of $21,694$ images. Frames (a total $11,058$ frames) from the remaining $14$ real and $14$ Deep Fake videos from the UADFV dataset and all images in the DARPA GAN set are used to test the SVM classifiers. We train SVM classifier with RBF kernels on the training data, with a grid search on the hyperparameters using $5$ fold cross validation.  



\begin{figure}[t]
	\centering
	\includegraphics[width=.95\linewidth]{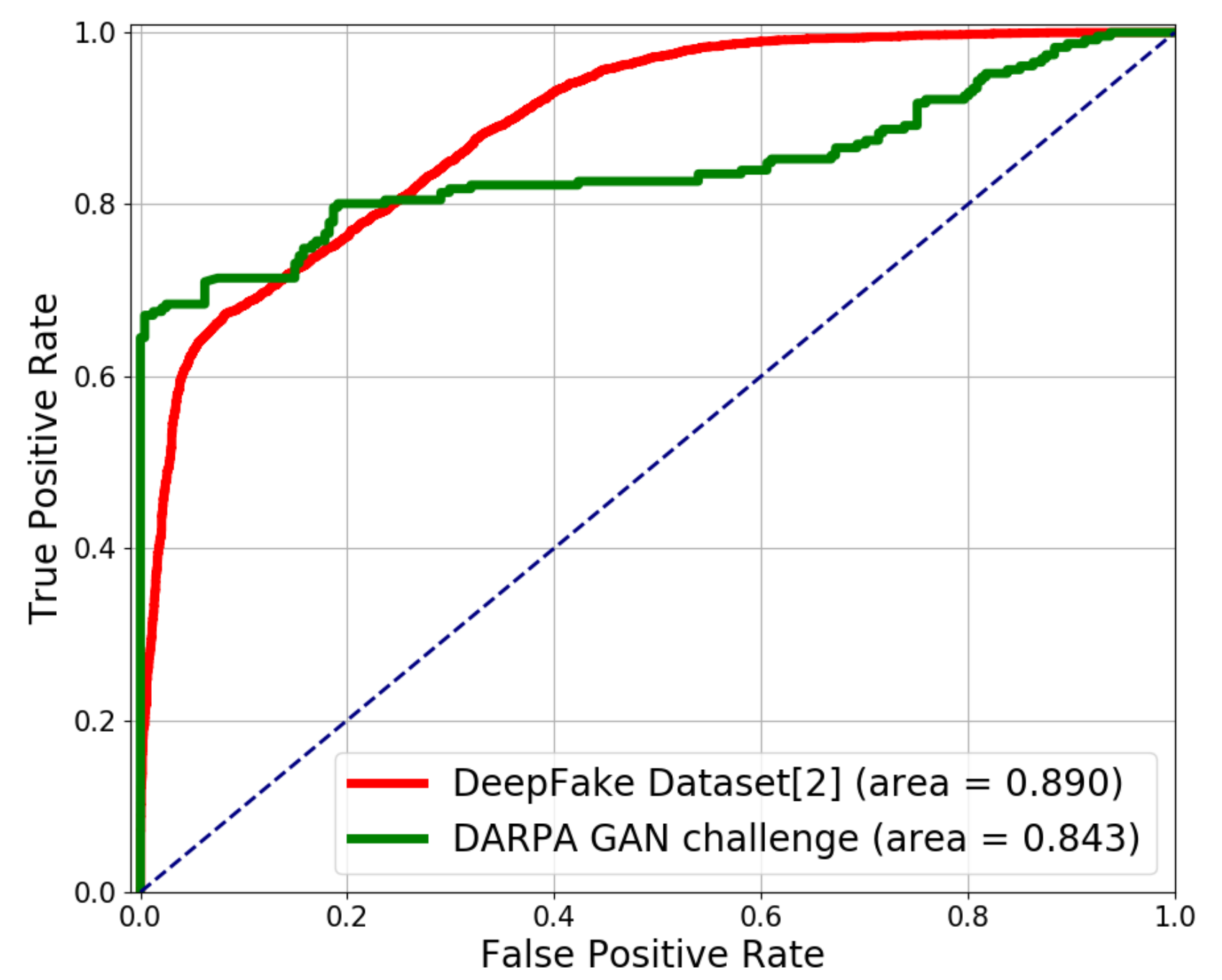}
	~\vspace{-1em}
	\caption{\em \small ROC curves of the SVM classification results, see texts for details.}
	\label{fig:img_auc}
	~\vspace{-2em}
\end{figure}

The performance, evaluated using individual frames as unit of analysis with Area Under ROC (AUROC) as the performance metric, is shown for the two datasets in Fig. \ref{fig:img_auc}. As these results show, on the UADFV dataset, the SVM classifier achieves an AUROC of $0.89$.  This indicates that the difference between head poses estimated from central region and whole face is a good feature to identify Deep Fake generated images. Similarly, on the DARPA GAN Challenge dataset, the AUROC of the SVM classifier is $0.843$. This results from the fact that the synthesized faces in the DARPA GAN challenges are often blurry, leading to difficulties to accurately predict facial landmark locations, and consequently the head pose estimations. We also estimate the performance using individual videos as unit of analysis for the UADFV dataset. This is achieved by averaging the classification prediction on frames over individual videos. The performance is shown in the last row of Table \ref{tab:Table1}. 



We also perform an ablation study to compare the performance of different types of features used in the SVM classifier. Specifically, we compare $five$ different types of features based on the rotation and translation of estimated 3D head pose in camera coordinates are also examined as in Table \ref{tab:Table1}. (1) As in Section \ref{inconsistency}, we simplified head poses as head orientations, $\vec{v}_a$ and $\vec{v}_c$. Classification using $\vec{v}_a - \vec{v}_c$ as features achieves $0.738$ AUROC on Deep Fake Dataset. This is expected, as this simplification neglects the translation and rotation on other axes. (2) As there are $3$ degrees of freedom in rotation, representing head pose rotation matrix as Rodrigues' rotation vector ($\vec{r}_a - \vec{r}_c$) could increase the AUROC to $0.798$. (3) Instead of Rodrigues' vector $\vec{r}\in R^{3}$, flatten the difference of 3 by 3 rotation matrices $R_a - R_c$ as features further improve the AUROC to $0.840$. (4) Introducing the difference of translation vectors $\vec{t}_a - \vec{t}_c$ to (1) and (2) results in AUROCs as $0.866$ and $0.890$, due to the increase of head poses in translation.

\begin{table}
\centering
\caption{AUROC based on videos and frames} 
\label{tab:Table1}
\begin{tabular}{lll} \toprule
features & frame & video \\ \midrule
$\vec{v}_a - \vec{v}_c$ & 0.738 & 0.888  \\
$\vec{r}_a - \vec{r}_c$ & 0.798 & 0.898  \\
$R_a - R_c$ & 0.853 & 0.913\\
$(\vec{v}_a - \vec{v}_c)\ \&\ (\vec{t}_a - \vec{t}_c)$& 0.840 & 0.949  \\
$(\vec{r}_a - \vec{r}_c)\ \&\ (\vec{t}_a - \vec{t}_c)$ & 0.866 & 0.954  \\
$(R_a - R_c)\ \&\ (\vec{t}_a - \vec{t}_c)$ & 0.890 & 0.974  \\ \bottomrule
\end{tabular}
\end{table}

\section{Conclusion}

In this paper, we propose a new method to expose AI-generated fake face images or videos (commonly known as the {\em Deep Fakes}). Our method is based on observations that such Deep Fakes are created by splicing synthesized face region into the original image, and in doing so, introducing errors that can be revealed when 3D head poses are estimated from the face images. We perform experiments to demonstrate this phenomenon and further develop a classification method based on this cue. We also report experimental evaluations of our methods on a set of real face images and Deep Fakes.

{\small
\bibliographystyle{IEEEbib}
\bibliography{ref}
}
\end{document}